\documentclass[10pt,twocolumn,letterpaper]{article}

\usepackage[datasets]{wacv}

\usepackage[utf8]{inputenc}
\usepackage{times}
\usepackage{epsfig}
\usepackage{graphicx}
\usepackage{amsmath,amssymb}
\usepackage{booktabs}
\usepackage{multirow}
\usepackage{xspace}
\usepackage{enumitem}
\usepackage{pifont}
\usepackage{color}
\usepackage{array}
\usepackage{float}
\usepackage{makecell}
\usepackage{tabularx}
\usepackage{stfloats}
\usepackage[pagebackref,breaklinks,colorlinks]{hyperref}

\newcommand{\method}{FairRSFM\xspace}

\def\wacvPaperID{1517}
\def\confName{WACV}
\def\confYear{2027}

\title{\method: A Biome-Aware Benchmark and Debiasing Framework for Remote Sensing Foundation Models}

\author{
Md Aminur Hossain\textsuperscript{1} \qquad
Omkumar Vaghasiya\textsuperscript{3} \qquad
Rajeev Ranjan Dwivedi\textsuperscript{3} \\
Vinod Kurmi\textsuperscript{3} \qquad
Biplab Banerjee\textsuperscript{2}
\\[2mm]
\textsuperscript{1}Space Applications Centre, ISRO, Ahmedabad, India\\
\textsuperscript{2}Centre of Studies in Resources Engineering (CSRE),
Indian Institute of Technology Bombay, India\\
\textsuperscript{3}Indian Institute of Science Education and Research Bhopal,
Bhopal, India\\
}

\begin{document}
\maketitle


\begin{abstract}
Remote sensing foundation models (RSFMs) are commonly evaluated using aggregate metrics, which can hide systematic performance disparities across ecological regions. We introduce \method, a biome-aware benchmark for evaluating ecological group robustness in RSFMs. \method maps georeferenced samples from 14 terrestrial biome classes into six ecologically meaningful macro-groups and evaluates models under a unified frozen-backbone evaluation protocol. The benchmark covers four downstream datasets: m-EuroSAT, m-BigEarthNet, m-SA-Crop-Type, and MMEarth20K with Dynamic World label maps. Using Prithvi-EO-2.0, SatMAE, and DOFA across three random seeds, we show that aggregate performance consistently masks biome-dependent disparities across architectures and tasks. For example, Prithvi-EO-2.0 reaches $90.98\%$ overall macro-F1 on m-EuroSAT but a mean worst-group score of only $83.72\%$, while m-SA-Crop-Type drops from $27.30\%$ overall mIoU to $18.47\%$ in the Xeric and Mineralogical group. We further evaluate Biome-Orthogonal Linear Probing (BOLP), Dynamic Biome Reweighting (DBR), and GroupDRO as complementary mitigation baselines. Their effectiveness is model- and task-dependent; for example, BOLP improves Prithvi-EO-2.0 worst-group F1@opt on m-BigEarthNet from $46.12\%$ to $50.27\%$ without updating the RSFM backbone. \method provides a reusable protocol for diagnosing and mitigating ecological robustness gaps in remote sensing foundation models. Code and datasets are available at: \url{https://github.com/aminurhossain/FairRSFM}.
\end{abstract}
\section{Introduction}

Remote sensing foundation models (RSFMs) have shown strong transferability across diverse Earth-observation tasks, including land-cover classification, crop mapping, flood segmentation, and semantic segmentation. Large pretrained encoders such as SatMAE~\cite{satmae}, Prithvi-EO-2.0~\cite{prithvi2}, DOFA~\cite{dofa}, SpectralGPT~\cite{spectralgpt}, and related RSFMs are increasingly used as general-purpose backbones for downstream remote-sensing applications. Benchmark suites such as GEO-Bench~\cite{geobench} have accelerated this trend by providing classification and segmentation tasks for systematic evaluation of pretrained models.

However, most RSFM evaluations still report only aggregate dataset-level metrics such as accuracy, macro-F1, or mean intersection-over-union (mIoU). These aggregate metrics answer whether a model performs well \emph{on average}, but they do not reveal whether the model performs consistently across ecological regions. This is a critical issue for remote sensing because Earth-observation data are spatially and ecologically structured: the appearance of the same semantic class can change substantially across forests, grasslands, deserts, Mediterranean ecosystems, boreal regions, and agricultural landscapes. A model may therefore achieve high average accuracy while underperforming in specific ecological regions that are visually, climatically, or geographically under-represented. This omission matters in practice because geospatial models can degrade substantially under realistic natural distribution shifts, including geographic, sensor, scale, and source changes~\cite{rolf2023evaluation,earthshift2026}.

We focus on \textbf{biome bias}, which we define as systematic disparity in RSFM downstream performance across biome-derived ecological groups. A biome is a large ecological region characterized by similar climate, vegetation, and environmental conditions. To measure this bias, we assign each georeferenced sample to a biome label using its geographic coordinates and an external terrestrial ecoregion reference map~\cite{terrestrialecoregions,ecoregions2017}. We then evaluate model performance separately within each biome-derived group rather than only reporting a single aggregate dataset-level score. This converts standard RSFM transfer evaluation into a group-robustness problem, where poor worst-group performance can indicate reduced robustness in specific ecological regimes.

Although the reference taxonomy contains 14 terrestrial biome classes, directly using all 14 classes as primary evaluation groups can be unstable for downstream datasets with limited or uneven biome coverage. We therefore adopt a two-level grouping strategy. Each sample is first mapped to one of the 14 raw terrestrial biome labels and then consolidated into six biome macro-groups reflecting broad spectral, phenological, hydrological, cryospheric, and surface-property regimes. These groups correspond to aseasonal high-biomass regions, high-amplitude phenological forests, transitional herbaceous and scrub landscapes, cryospheric and short-cycle regions, xeric and mineralogical surfaces, and hydrologically modulated ecosystems. The macro-groups are intended to reduce sparsity by pooling under-represented raw biomes while retaining ecologically meaningful distinctions; the original 14 labels are preserved for finer-grained analysis when sufficient samples are available.

We introduce \method, a biome-aware benchmark and debiasing framework for evaluating ecological group robustness in remote sensing foundation models. Our benchmark design is inspired by GEO-Bench but differs in objective. GEO-Bench was designed to evaluate foundation models across a diverse set of Earth-monitoring tasks, including classification and segmentation datasets~\cite{geobench}. In contrast, \method uses three selected GEO-Bench tasks together with an MMEarth20K Dynamic World segmentation subset to ask a different question: \emph{Do RSFMs generalize consistently across biome-derived ecological groups?} We instantiate this protocol with three complementary RSFMs---Prithvi-EO-2.0, SatMAE, and DOFA---under the same frozen-backbone evaluation setting. This question is particularly important for global Earth-observation deployment, where a model trained or evaluated mostly on dominant ecological regimes may underperform in less represented regions.

Beyond diagnosis, \method includes biome-aware mitigation baselines under the
same frozen-backbone downstream evaluation protocol. We evaluate standard Empirical Risk Minimization (ERM) together with Dynamic Biome Reweighting with Validation Feedback (DBR), Biome-Orthogonal Linear Probing (BOLP), and GroupDRO~\cite{groupdro}. BOLP removes biome-associated directions from frozen RSFM embeddings before training the task head, providing a representation-level mitigation strategy. This allows us to analyze whether biome-dependent disparities can be reduced without full backbone fine-tuning.

Our main contributions are as follows:
\begin{itemize}[leftmargin=*]
    \item We introduce \method, a biome-aware benchmark and debiasing framework for studying ecological group robustness in remote sensing foundation models across four classification and segmentation datasets.

    \item We propose a biome-labeling pipeline that first assigns georeferenced samples to the 14 terrestrial biome classes and then groups them into six biome macro-groups reflecting spectral, phenological, hydrological, cryospheric, and surface-property regimes.

    \item We evaluate Prithvi-EO-2.0, SatMAE, and DOFA using a unified frozen-backbone downstream evaluation protocol over three random seeds, and report both standard task metrics and biome-aware robustness measures, showing that strong aggregate performance can hide substantial biome-dependent disparities.

    \item We evaluate complementary mitigation baselines, including DBR, BOLP, and GroupDRO, showing that improvements in worst-group robustness are model- and task-dependent and can involve trade-offs with aggregate performance.
\end{itemize}

\section{Related Work}

\noindent\textbf{Remote sensing foundation models.}
RSFMs learn transferable representations from large-scale Earth observation through self-supervised pretraining on multispectral, temporal, and multimodal imagery~\cite{rsfm_survey_multimodal}. Representative approaches include masked-autoencoding models such as SatMAE~\cite{satmae}, Scale-MAE~\cite{scalemae}, and SatMAE++~\cite{satmaepp}; contrastive and hybrid methods such as CROMA~\cite{croma} and SoftCon~\cite{softcon}; and broad-coverage models including Prithvi-EO-2.0~\cite{prithvi2}, DOFA~\cite{dofa}, SpectralGPT~\cite{spectralgpt}, and SkySense~\cite{skysense}. These models are typically evaluated using aggregate task metrics, which do not reveal whether performance remains consistent across ecological regions; \method targets this complementary robustness dimension.

\smallskip
\noindent\textbf{Remote sensing benchmarks.}
Standardized suites have driven RSFM evaluation. GEO-Bench~\cite{geobench} provides harmonized classification and segmentation tasks, from which we use m-EuroSAT, m-BigEarthNet, and m-SA-Crop-Type; we additionally include MMEarth20K, derived from MMEarth~\cite{mmearth} with Dynamic World~\cite{dynamicworld} label maps, for dense land-cover evaluation. PANGAEA~\cite{pangaea} broadens geospatial foundation-model assessment
across tasks, resolutions, and geographies, while REOBench evaluates
robustness to image corruptions~\cite{reobench}. Recent work has also quantified remote-sensing dataset bias through size, angle, class, density, and scene factors using diversity- and evenness-based indices~\cite{rsd_biaseval}. Existing benchmarks primarily summarize performance at the dataset level, whereas \method complements them by measuring how RSFM performance varies across biome-derived ecological groups.

\begin{figure*}
    \centering
    \includegraphics[width=0.95\linewidth]{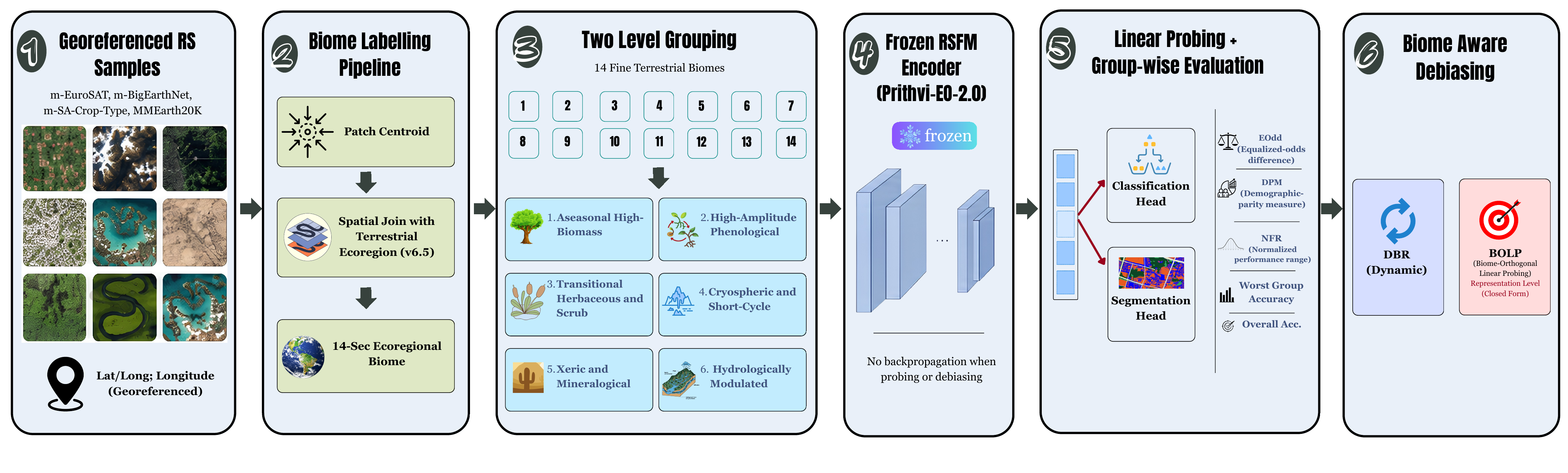}
    \caption{Overview of \method. Georeferenced samples are mapped from 14 terrestrial biomes to six macro-groups and evaluated under a frozen-backbone downstream evaluation protocol. BOLP removes biome-associated directions from frozen RSFM embeddings before training the task head.}
    \label{fig:main}
\end{figure*}

\smallskip
\noindent\textbf{Group robustness and representation debiasing.}
Group-aware evaluation compares performance across subgroups rather than only population-level averages~\cite{fairnesssurvey}, motivating worst-group and GroupDRO-style objectives~\cite{groupdro,wilds}. In remote sensing, biomes provide globally defined and ecologically interpretable analysis groups linked to climate, vegetation, and land-cover appearance, making them suitable for subgroup robustness analysis under distribution shift~\cite{hardt2016equality,woodworth2017learning,shao2024fairnessshift}. Parameter-efficient adaptation of RSFMs has been widely studied~\cite{yin2025remote}, while debLoRA~\cite{deblora} specifically addresses debiasing under long-tailed remote-sensing data. More general representation-debiasing methods remove attribute-associated directions from embeddings~\cite{bolukbasi2016man,ravfogel2020null}. Our Biome-Orthogonal Linear Probing (BOLP) extends this principle to ecological robustness in RSFMs by identifying and removing biome-associated directions from frozen embeddings through a closed-form projection, without backbone fine-tuning or additional trainable parameters.

\section{FairRSFM Benchmark}

\subsection{Task Families and Datasets}

\method contains two application families: \textbf{classification} and \textbf{semantic segmentation}. Figure~\ref{fig:main} summarizes the overall \method benchmark and mitigation pipeline. Classification tasks evaluate whether frozen RSFM features are discriminative for patch-level labels, while segmentation tasks evaluate whether RSFM representations support dense prediction under biome-derived ecological variation. Table~\ref{tab:datasets} summarizes the four-dataset benchmark suite. We use three GEO-Bench tasks covering single-label classification, multi-label classification, and crop-type segmentation, and add an MMEarth20K subset with Dynamic World label maps to increase dense land-cover segmentation diversity.

\begin{table*}[t]
\centering
\caption{\method dataset suite. GEO-Bench datasets are selected for classification and segmentation; the MMEarth20K subset increases dense land-cover segmentation diversity. Sample counts for GEO-Bench tasks follow the modified benchmark setting.}
\label{tab:datasets}
\scalebox{0.9}{
\begin{tabular}{l l c c c c l}
\toprule
Dataset & Task & Classes & Train & Val & Test & Notes \\
\midrule
m-EuroSAT~\cite{geobench} & Classification & 10 & 2,000 & 1,000 & 1,000 & Sentinel-2, land cover \\
m-BigEarthNet~\cite{geobench,bigearthnet} & Multi-label classification & 43 & 20,000 & 1,000 & 1,000 & Sentinel-2, multi-label land cover \\
m-SA-Crop-Type~\cite{geobench} & Segmentation & 10 & 3,000 & 1,000 & 1,000 & Sentinel-2, crop type \\
MMEarth20K~\cite{mmearth,dynamicworld} & Segmentation & 9 & 16,000 & 2,000 & 2,000 & Dynamic World label maps \\
\bottomrule
\end{tabular}
}
\end{table*}

For MMEarth20K, we construct a 20,000-sample subset with 10\,m Dynamic World label maps~\cite{dynamicworld} from the global MMEarth dataset~\cite{mmearth}. To stress-test ecological robustness under realistic long-tails and evaluate whether debiasing prevents minority-group collapse, macro-group quotas are derived via inverse-frequency weighting of the parent MMEarth distribution ($28.0\%$ to $6.0\%$) and allocated uniformly across constituent 14 RESOLVE biomes, with tiles sampled randomly without class constraints (16,000 train, 2,000 val, 2,000 test; details in Supp.~S4).

Together, these four datasets provide public geocoordinates for reproducible biome assignment, span classification and dense segmentation, and exhibit measurable ecological imbalance. The lightweight metadata join makes \method readily extensible to other georeferenced remote-sensing datasets.

\subsection{Biome Label Construction}

The central grouping variable in \method is derived from terrestrial biome information. We use the 14-biome taxonomy from global terrestrial ecoregion products~\cite{terrestrialecoregions,ecoregions2017}. For each georeferenced sample, we compute the patch centroid and spatially join it with an external biome reference map. The resulting raw biome ID is stored as image-level metadata. For segmentation datasets, the entire patch is assigned a biome label based on its centroid, which keeps the grouping protocol consistent across classification and dense prediction tasks. This can be noisy for ecotone or multi-biome tiles; we treat it as a consistent grouping rule rather than a pixel-perfect ecological map (see Limitations).

Directly using all 14 raw biome classes as primary evaluation groups can be unstable for downstream datasets with limited or uneven biome coverage. Therefore, \method uses a two-level grouping strategy. Each sample is first assigned a raw biome label \(b_i\in\{1,\ldots,14\}\), and the raw label is then mapped to a biome macro-group \(g_i\in\{1,\ldots,6\}\). The six macro-groups are designed to reflect broad spectral, phenological, hydrological, cryospheric, and surface-property regimes. The mapping is fixed across all datasets, models, and mitigation methods. They are used as the primary groups for ecological robustness evaluation and mitigation, while the original 14 biome labels are retained for fine-grained analysis when sufficient samples are available. Supplementary Fig.~S1 shows the global spatial distribution of these macro-groups.

Unknown, water-only, or unmatched samples are excluded from the primary
group-based summaries unless explicitly stated. Some datasets cover only a
subset of the six macro-groups; therefore, all group-robustness metrics are
computed over the matched macro-groups present in each dataset split.

\subsection{Biome Metadata Schema}

Each sample is associated with the following metadata:
\((x_i, y_i, d_i, \ell_i, b_i, g_i)\),
where \(x_i\) is the input image, \(y_i\) is the class label or segmentation mask, \(d_i\) is the dataset name, \(\ell_i=(\mathrm{lat}_i,\mathrm{lon}_i)\) is the sample location, \(b_i\in\{1,\ldots,14\}\) is the raw terrestrial biome label, and \(g_i\in\{1,\ldots,6\}\) is the derived biome macro-group label. The macro-group is obtained through a deterministic mapping \(g_i=\psi(b_i)\), where \(\psi\) is defined in Supplementary Table~S1. For single-label classification, \(y_i\in\{1,\ldots,C\}\); for multi-label classification, \(y_i\in\{0,1\}^{C}\); and for segmentation, \(y_i \in \{1,\ldots,C\}^{H\times W}\).

The raw biome label \(b_i\) enables fine-grained ecological analysis, while the macro-group label \(g_i\) is used for the primary group-robustness metrics and mitigation methods. This separation allows \method to preserve ecological detail without making the primary robustness analysis overly sensitive to sparse biome classes.

\subsection{Evaluation Metrics}

Let $\mathcal{G}_d$ be the set of active biome macro-groups present in dataset $d$ ($K = |\mathcal{G}_d|$), and let $M_g$ denote the task-specific metric evaluated on group $g \in \mathcal{G}_d$. To diagnose ecological robustness, calibration, and cross-group disparity, \method uses three complementary metric tiers:

\smallskip
\noindent\textbf{1. Per-Task Performance Metrics.}
For single-label classification (m-EuroSAT), we report macro-F1 over present classes $C_g$:
$\mathrm{Macro\text{-}F1}_g =
\frac{1}{|C_g|}\sum_{c\in C_g}\mathrm{F1}_{g,c}$.
For multi-label classification (m-BigEarthNet), we report macro-F1 at per-class optimal validation thresholds $\tau_c \in [0.1, 0.95]$, denoted
$\mathrm{F1@opt}_g =
\frac{1}{|C_g|}\sum_{c\in C_g}\mathrm{F1}_{g,c}(\tau_c)$.
For segmentation (m-SA-Crop-Type, MMEarth20K), we report group-wise mean Intersection-over-Union:
\begin{equation}
\mathrm{mIoU}_g =
\frac{1}{|C_g|}
\sum_{c\in C_g}
\frac{\mathrm{TP}_{g,c}}
{\mathrm{TP}_{g,c}+\mathrm{FP}_{g,c}+\mathrm{FN}_{g,c}}.
\end{equation}

\smallskip
\noindent\textbf{2. Group Disparity \& Robustness Summaries.}
From the group performance scores $\{M_g\}_{g\in\mathcal{G}_d}$ and their mean
$\bar{M}=\frac{1}{K}\sum_g M_g$, we report:

\noindent\textbf{Worst-Group Score ($M_{\mathrm{worst}}$):}
$M_{\mathrm{worst}}=\min_{g\in\mathcal{G}_d}M_g$. $M_{\mathrm{worst}}$ is computed per-seed then aggregated, not as min of the group-level means.

\noindent\textbf{Normalized Failure Range ($\mathrm{NFR}\downarrow$):}
$\mathrm{NFR}=
\frac{\max_g M_g-\min_g M_g}{\bar{M}+\epsilon}$,
measuring the relative cross-group performance spread.

For datasets containing only two matched macro-groups, we emphasize $M_{\mathrm{worst}}$ and interpret range-based summaries cautiously.

\smallskip
\noindent\textbf{3. Calibration \& Ecological Group-Disparity Metrics.}

\noindent\textbf{Expected Calibration Error ($\mathrm{ECE} \downarrow$):}
Model confidence scores are partitioned into $B=10$ equal-width bins $\{B_m\}_{m=1}^{B}$:
\begin{equation}
\mathrm{ECE} =
\sum_{m=1}^{B}
\frac{|B_m|}{N}
\left|
\mathrm{acc}(B_m)-\mathrm{conf}(B_m)
\right|.
\end{equation}

\noindent\textbf{Equalized-Odds-Style Disparity ($\mathrm{EOdd} \downarrow$):}
Evaluates error-rate differences across groups via Youden's $J$-index, following equalized-odds-style group comparisons~\cite{hardt2016equality,fairnesssurvey}.
$J_{g,c}=\mathrm{TPR}_{g,c}-\mathrm{FPR}_{g,c}$.
The metric measures the maximum cross-group disparity averaged across classes:
\begin{equation}
\mathrm{EOdd} =
\max_{g,g'\in\mathcal{G}_d}
\frac{1}{|C|}
\sum_{c\in C}
\left|
J_{g,c}-J_{g',c}
\right|.
\end{equation}

\noindent\textbf{Demographic Parity Metric ($\mathrm{DPM} \downarrow$):}
Quantifies differences in positive prediction rates
$\mathrm{PPR}_{g,c}=P(\hat{y}_c=1\mid g)$
across ecological strata:
\begin{equation}
\mathrm{DPM} =
\max_{g,g'\in\mathcal{G}_d}
\frac{1}{|C|}
\sum_{c\in C}
\left|
\mathrm{PPR}_{g,c}-\mathrm{PPR}_{g',c}
\right|.
\end{equation}

EOdd and DPM are used solely as descriptive group-disparity summaries
under ecological shift, not as claims about protected human attributes.
For readability, ECE, NFR, EOdd, and DPM are reported as percentages
($\times 100$) in all result tables. For segmentation, ECE, EOdd, and DPM are computed over valid pixels and
aggregated across classes using the definitions above.

\section{Biome-Aware Debiasing Framework}

Let \(\mathcal{D}_{\mathrm{tr}}=\{(x_i,y_i,g_i)\}_{i=1}^{N}\) be the training set with input image \(x_i\), task label \(y_i\), and biome macro-group label \(g_i\in\mathcal{G}\) (see Supplementary Table~S1). Let \(\phi:X\rightarrow\mathbb{R}^{d}\) be a frozen RSFM encoder and \(z_i=\phi(x_i)\) the corresponding representation. All strategies below keep \(\phi\) fixed and train only the downstream head \(h_\theta\), isolating the effect of the mitigation strategy from backbone fine-tuning. The task loss $\ell$ is cross-entropy for single-label classification,
binary cross-entropy for multi-label classification, and pixel-wise
cross-entropy for segmentation.

\subsection{Dynamic Biome Reweighting with Validation Feedback}

We use Dynamic Biome Reweighting with Validation Feedback (DBR) as a loss-level mitigation strategy. DBR initializes group weights using inverse frequency and adapts them during training according to per-group validation performance.

Let \(p_g=n_g/N\) be the empirical frequency of biome macro-group \(g\). DBR initializes the normalized weights as
\begin{equation}
w_g^{(0)}=
\frac{1/p_g}
{\frac{1}{|\mathcal{G}|}\sum_{g'\in\mathcal{G}}1/p_{g'}}
\label{eq:wg}.
\end{equation}

Frequency alone does not necessarily identify the worst-performing group. Therefore, every \(T_{\mathrm{upd}}\) epochs, the current model is evaluated separately on each biome macro-group. Let \(M_g^{(t)}\) denote the task-specific validation metric for group \(g\)
at update step \(t\), and let
\(\bar{M}^{(t)}=\frac{1}{|\mathcal{G}|}\sum_g M_g^{(t)}\).
DBR increases the weight of groups performing below the mean and decreases it for groups performing above the mean, followed by normalization to preserve unit average:
\begin{align}
\tilde{w}^{(t+1)}_g
&=
w^{(t)}_g
\left(
1+\alpha
\frac{\bar{M}^{(t)}-M_g^{(t)}}
{\bar{M}^{(t)}+\epsilon}
\right),\\
w^{(t+1)}_g
&=
\frac{\tilde{w}^{(t+1)}_g}
{\frac{1}{|\mathcal{G}|}\sum_{g'}\tilde{w}^{(t+1)}_{g'}}.
\end{align}
Here, \(\alpha\) controls feedback strength and \(\epsilon\) ensures numerical stability. The training objective uses the current group weights:
\begin{equation}
\mathcal{L}_{\mathrm{DBR}}
=
\frac{1}{N}
\sum_{i=1}^{N}
w^{(t)}_{g_i}\,
\ell(h_\theta(\phi(x_i)),y_i).
\end{equation}
DBR reuses validation metrics already computed for group-robustness evaluation, introduces no additional trainable parameters, and applies to single-label classification, multi-label classification, and segmentation.

\subsection{Biome-Orthogonal Linear Probing}

Biome-Orthogonal Linear Probing (BOLP), our representation-level mitigation method, operates directly on the geometry of frozen RSFM embeddings. It is a closed-form module inserted between the frozen encoder and downstream head that removes dominant biome-associated directions without updating the backbone.

\textbf{Precomputation.}
For each group, we compute the centroid
\(\mu_g=\frac{1}{n_g}\sum_{i:g_i=g}z_i\),
their mean
\(\mu_0=\frac{1}{|\mathcal{G}|}\sum_g\mu_g\),
and the centered group-mean matrix
\begin{equation}
B=
\left[
\mu_1-\mu_0,\ldots,\mu_{|\mathcal{G}|}-\mu_0
\right]
\in\mathbb{R}^{d\times|\mathcal{G}|}.
\end{equation}
From the thin SVD \(B=U\Sigma V^\top\), the top \(k\) left singular vectors
\(U_k\in\mathbb{R}^{d\times k}\) span the dominant inter-group mean-difference directions. Because the columns of \(B\) sum to zero, at most \(|\mathcal{G}|-1\) directions are non-trivial; hence \(k\leq|\mathcal{G}|-1\). The projector onto their orthogonal complement is
\(P_k=I_d-U_kU_k^\top\).
The buffers \(\{\mu_g\}\), \(\mu_0\), and \(P_k\) are computed once from the training set and then fixed.

\textbf{Debiasing transform.}
For a sample with group label \(g_i\), BOLP removes the first-order group-mean shift and projects out the dominant inter-group mean-difference subspace:
\begin{equation}
\hat{z}_i=
P_k\big(z_i-\mu_{g_i}+\mu_0\big).
\end{equation}
Thus, \(\hat{z}_i\) is recentered to a common reference and restricted to the orthogonal complement of the selected biome-associated directions.

\textbf{Linear probing.}
The downstream head is trained on the transformed features using
\[
\mathcal{L}_{\mathrm{BOLP}}
=
\frac{1}{N}\sum_i
\ell(h_\theta(\hat{z}_i),y_i),
\]
with only \(\theta\) trainable. At inference, the same transform is applied when the biome group is available; for georeferenced remote-sensing data, it can be obtained using the same coordinate-to-biome lookup employed during preprocessing. If the group label is unavailable, group-specific residualization is skipped and only the global projector \(P_k\) is applied. BOLP adapts representation debiasing and null-space projection~\cite{bolukbasi2016man,ravfogel2020null} to ecological robustness in frozen RSFM embeddings, adding no trainable parameters and leaving the backbone unchanged.

\subsection{Group Distributionally Robust Optimization}

As a stronger loss-level baseline from the group-robustness literature, we include GroupDRO~\cite{groupdro} under the same frozen-backbone protocol. GroupDRO maintains adaptive biome-group weights and upweights high-loss macro-groups while training only the downstream head, leaving \(\phi\) frozen. Together, ERM, DBR, BOLP, and GroupDRO provide complementary unweighted, validation-feedback, representation-level, and worst-group-oriented strategies under the same controlled protocol.

\smallskip
\noindent\textbf{Method Comparison.}
ERM uses the standard unweighted task loss. DBR and GroupDRO operate at the loss level by adapting biome-group weights using validation feedback or group loss, respectively. BOLP instead intervenes in the frozen representation space by recentering group means and projecting out dominant inter-group directions. All strategies retain the full training set and introduce no trainable parameters beyond the downstream task head.

\section{Experiments}

\noindent \textbf{Experimental Protocol and Pretrained Baselines.}
To evaluate ecological robustness across diverse remote sensing representation paradigms, we benchmark three foundational architectures using their official pretraining configurations and checkpoints:
(i)~\textbf{Prithvi-EO-2.0}~\cite{prithvi2} (300M spatio-temporal ViT) using its official 6-band HLS configuration and channel statistics;
(ii)~\textbf{SatMAE}~\cite{satmae} (ViT-Large, 304M parameters) using its official 10 Sentinel-2 bands and pretraining statistics; and
(iii)~\textbf{DOFA}~\cite{dofa} (ViT-Base, 86M parameters) with dynamic continuous wavelength-conditioned positional embeddings ($0.44\mu\text{m}\text{--}2.20\mu\text{m}$).
Dataset splits and task definitions follow \textbf{GEO-Bench}~\cite{geobench}
for m-EuroSAT, m-BigEarthNet, and m-SA-Crop-Type. For MMEarth20K, we use
the 20,000-sample subset and train/validation/test split defined in
Section~3.1.

\smallskip
\noindent \textbf{Controlled Frozen-Encoder Evaluation.}
In contrast to the original model papers that frequently fine-tuned backbones end-to-end for dense prediction (e.g., in DOFA~\cite{dofa} and SatMAE~\cite{satmae}), \method enforces a \textbf{strictly frozen encoder} $\phi(x)$ across all classification and dense segmentation tasks. Backpropagation updates only the task-specific classification head or
the segmentation decoder listed in Table~\ref{tab:training_config}. This prevents backbone adaptation and enables controlled within-backbone
comparisons of ERM, BOLP, DBR, and GroupDRO. Because the downstream heads
follow model-specific probe/decoder implementations
(Table~\ref{tab:training_config}), absolute cross-model scores are interpreted
descriptively, while mitigation effects are compared within each frozen backbone.

\smallskip

\noindent\textbf{Training and Optimization Details.}
Table~\ref{tab:training_config} summarizes the configurations. All downstream
heads and decoders are optimized on $224\times224$ inputs using AdamW
($\beta_1=0.9$, $\beta_2=0.999$, base $\mathrm{lr}=10^{-3}$ cosine-annealed
to $10^{-6}$, weight decay $10^{-4}$) for up to 50 epochs with patience 15.
DBR uses $\alpha=0.5$, $T_{\mathrm{upd}}=1$, and $\epsilon=10^{-4}$;
BOLP uses $k=|\mathcal{G}_d|-1$ ($k=2,2,1,5$ for the four datasets,
respectively). Hyperparameters are selected on validation data, and test
results are reported over three seeds. Sensitivity analyses are provided in
the Supplementary Material.

\begin{table}[t]
\centering
\caption{Unified benchmarking configurations across foundation models.}
\label{tab:training_config}
\setlength{\tabcolsep}{2.5pt}
\renewcommand{\arraystretch}{1.05}
\scriptsize
\resizebox{\linewidth}{!}{%
\begin{tabular}{lccc}
\toprule
\textbf{Configuration} & \textbf{Prithvi-EO-2.0}~\cite{prithvi2} & \textbf{SatMAE}~\cite{satmae} & \textbf{DOFA}~\cite{dofa} \\
\midrule
Backbone & Spatio-Temporal ViT (300M) & ViT-Large (304M) & ViT-Base (86M) \\
Embedding Dim & 1024 ($16\times16$ patch) & 1024 ($16\times16$ patch) & 768 ($16\times16$ patch) \\
Input Bands & 6 HLS Bands & 10 Sentinel-2 Bands & 9--12 Wavelength Bands \\
Encoder Status & \textbf{Frozen} (0 updates) & \textbf{Frozen} (0 updates) & \textbf{Frozen} (0 updates) \\
Classification Head & Linear / MLP probe & Linear probe & BatchNorm1d + Linear \\
Segmentation Head & FCN decoder & Conv decoder & Multi-level UPerNet~\cite{dofa} \\
Optimizer & AdamW (Cosine LR) & AdamW (Cosine LR) & AdamW (Cosine LR) \\
Image Size & $224\times224$ & $224\times224$ & $224\times224$ \\
Batch Size & 64 (Cls) / 32 (Seg) & 64 (Cls) / 32 (Seg) & 64 (Cls) / 32 (Seg) \\
Epochs & Up to 50 (Patience=15) & Up to 50 (Patience=15) & Up to 50 (Patience=15) \\
Random Seeds & 3 independent runs & 3 independent runs & 3 independent runs \\
\bottomrule
\end{tabular}}
\end{table}

\subsection{Biome Distribution Analysis}

Before analyzing performance, we first examine the biome composition of the evaluation splits. Table~\ref{tab:ng_all_splits} reports per-macro-group counts across dataset splits. Matched-group coverage is 98.8\%, 88.5\%, 99.9\%, and 100\% for
m-EuroSAT, m-BigEarthNet, m-SA-Crop-Type, and MMEarth20K, respectively. Coverage is uneven: m-EuroSAT and m-BigEarthNet have only three matched macro-groups, with the smallest matched-group test sizes of $N_g=25$ and $N_g=14$; m-SA-Crop-Type has two groups; MMEarth20K covers all six. This imbalance, including the well-known concentration of m-BigEarthNet in temperate and Mediterranean biomes~\cite{bigearthnetmm}, motivates biome-aware evaluation rather than only aggregate dataset-level metrics.

\begin{table}[t]
\centering
\caption{Per-macro-group sample counts across Train/Validation/Test splits for all benchmark datasets. Empty cells indicate absent groups. Unknown or unmatched samples are excluded from worst-group scores.}
\label{tab:ng_all_splits}
\setlength{\tabcolsep}{3pt}
\scriptsize
\resizebox{\linewidth}{!}{%
\begin{tabular}{llccccccc}
\toprule
\textbf{Dataset} & \textbf{Split} & \textbf{Pheno.} & \textbf{Biomass} & \textbf{Trans.} & \textbf{Cryo.} & \textbf{Hydro.} & \textbf{Xeric} & \textbf{Unk.} \\
\midrule
\multirow{3}{*}{m-EuroSAT} 
 & Train & 1,430 & 50 & 481 & -- & -- & -- & 39 \\
 & Valid & 741 & 15 & 227 & -- & -- & -- & 17 \\
 & Test & 724 & 25 & 239 & -- & -- & -- & 12 \\
\midrule
\multirow{3}{*}{m-BigEarthNet} 
 & Train & 11,269 & 356 & 6,291 & -- & -- & -- & 2,084 \\
 & Valid & 518 & 17 & 335 & -- & -- & -- & 130 \\
 & Test & 535 & 14 & 336 & -- & -- & -- & 115 \\
\midrule
\multirow{3}{*}{m-SA-Crop-Type} 
 & Train & -- & -- & 2,640 & -- & -- & 358 & 2 \\
 & Valid & -- & -- & 840 & -- & -- & 158 & 2 \\
 & Test & -- & -- & 875 & -- & -- & 124 & 1 \\
\midrule
\multirow{3}{*}{MMEarth20K} 
 & Train & 4,078 & 4,478 & 2,883 & 2,216 & 964 & 1,381 & -- \\
 & Valid & 519 & 553 & 375 & 294 & 123 & 136 & -- \\
 & Test & 503 & 569 & 342 & 290 & 113 & 183 & -- \\
\bottomrule
\end{tabular}%
}
\end{table}

\noindent \textbf{Global and Group Performance Overview}
Table~\ref{tab:results_prithvi_mean_std}, Table~\ref{tab:results_satmae_mean_std}, and Table~\ref{tab:results_dofa_mean_std} summarize the downstream test performance of Prithvi-EO-2.0, SatMAE, and DOFA across all four datasets. The models achieve strong aggregate performance on classification and varied dense-segmentation performance, yet exhibit substantial ecological disparities under standard ERM.

\begin{table}[t]
\centering

\caption{Prithvi-EO-2.0: Mean and Std over seeds across datasets.}
\label{tab:results_prithvi_mean_std}
\setlength{\tabcolsep}{1.5pt}
\renewcommand{\arraystretch}{1.0}
\scriptsize
\resizebox{\linewidth}{!}{%
\begin{tabular}{llcccccc}
\toprule
\textbf{Dataset} & \textbf{Method} & \textbf{Test metric} & \textbf{$M_{\text{worst}}$} & \textbf{ECE $\downarrow$} & \textbf{NFR $\downarrow$} & \textbf{EOdd $\downarrow$} & \textbf{DPM $\downarrow$} \\
\midrule
\multirow{4}{*}{m-EuroSAT} 
 & ERM & 90.98$\pm$2.36 & 83.72$\pm$0.87 & 4.02$\pm$1.64 & \underline{5.96$\pm$3.28} & 20.50$\pm$1.75 & \underline{9.91$\pm$0.21} \\
 & \textbf{BOLP} & \underline{93.42$\pm$0.35} & \underline{84.34$\pm$0.00} & \textbf{2.37$\pm$0.04} & 9.82$\pm$0.30 & \textbf{17.06$\pm$0.16} & \textbf{9.61$\pm$0.03} \\
 & \underline{DBR} & 91.36$\pm$1.01 & \textbf{84.35$\pm$0.50} & 5.50$\pm$2.29 & \textbf{5.85$\pm$0.70} & 21.23$\pm$1.39 & 10.90$\pm$0.17 \\
 & GroupDRO & \textbf{94.14$\pm$0.47} & 84.34$\pm$0.00 & \underline{3.17$\pm$0.40} & 10.34$\pm$0.51 & \underline{17.31$\pm$0.12} & 10.04$\pm$0.10 \\
\midrule
\multirow{4}{*}{m-BigEarthNet} 
 & ERM & \underline{60.09$\pm$0.99} & 46.12$\pm$1.51 & 2.12$\pm$0.28 & 31.14$\pm$1.99 & \textbf{42.93$\pm$1.90} & \textbf{12.66$\pm$0.27} \\
 & \textbf{BOLP} & \textbf{62.75$\pm$0.16} & \textbf{50.27$\pm$0.63} & \underline{2.07$\pm$0.08} & \underline{20.92$\pm$0.39} & \underline{45.93$\pm$0.76} & \underline{13.29$\pm$0.11} \\
 & DBR & 58.05$\pm$0.32 & \underline{48.04$\pm$0.63} & 3.70$\pm$0.13 & 29.43$\pm$1.45 & 49.34$\pm$0.51 & 13.40$\pm$0.40 \\
 & \underline{GroupDRO} & 54.42$\pm$0.60 & 46.82$\pm$0.60 & \textbf{1.71$\pm$0.00} & \textbf{19.19$\pm$8.16} & 46.99$\pm$3.03 & 15.34$\pm$0.03 \\
\midrule
\multirow{4}{*}{MMEarth20K} 
 & ERM & \textbf{47.99$\pm$0.32} & 33.75$\pm$0.58 & 3.47$\pm$0.29 & 28.19$\pm$0.64 & \underline{22.51$\pm$2.36} & 9.78$\pm$0.19 \\
 & BOLP & 43.31$\pm$0.85 & 30.64$\pm$1.20 & 2.76$\pm$0.22 & 30.66$\pm$3.17 & \textbf{19.46$\pm$1.67} & \textbf{8.13$\pm$0.53} \\
 & \underline{DBR} & \underline{47.84$\pm$0.22} & \underline{34.76$\pm$0.42} & \underline{2.14$\pm$0.13} & \underline{26.42$\pm$0.62} & 23.15$\pm$3.22 & \underline{9.73$\pm$0.12} \\
 & \textbf{GroupDRO} & 46.99$\pm$0.62 & \textbf{37.16$\pm$0.26} & \textbf{1.69$\pm$0.08} & \textbf{15.34$\pm$1.91} & 24.45$\pm$0.20 & 10.50$\pm$0.15 \\
\midrule
\multirow{4}{*}{m-SA-Crop-Type} 
 & ERM & 27.30$\pm$0.15 & 18.47$\pm$0.16 & \underline{2.98$\pm$0.18} & 35.31$\pm$0.54 & 24.85$\pm$0.62 & 7.68$\pm$0.25 \\
 & \underline{BOLP} & \textbf{28.27$\pm$0.21} & \underline{19.38$\pm$0.19} & 5.94$\pm$1.18 & \underline{29.95$\pm$2.08} & 22.09$\pm$0.66 & 7.01$\pm$0.31 \\
 & \textbf{DBR} & 27.18$\pm$1.86 & \textbf{19.81$\pm$0.42} & \textbf{2.84$\pm$0.09} & \textbf{28.45$\pm$5.10} & \underline{20.31$\pm$2.42} & \underline{6.43$\pm$0.27} \\
 & GroupDRO & \underline{28.26$\pm$0.14} & 19.09$\pm$0.26 & 6.42$\pm$0.20 & 39.63$\pm$0.94 & \textbf{16.91$\pm$0.64} & \textbf{6.19$\pm$0.01} \\
\bottomrule
\end{tabular}
}

\vspace{0.4em}

\caption{SatMAE: Mean and Std over seeds across datasets.}
\label{tab:results_satmae_mean_std}
\setlength{\tabcolsep}{1.5pt}
\renewcommand{\arraystretch}{1.0}
\scriptsize
\resizebox{\linewidth}{!}{%
\begin{tabular}{llcccccc}
\toprule
\textbf{Dataset} & \textbf{Method} & \textbf{Test metric} & \textbf{$M_{\text{worst}}$} & \textbf{ECE $\downarrow$} & \textbf{NFR $\downarrow$} & \textbf{EOdd $\downarrow$} & \textbf{DPM $\downarrow$} \\
\midrule
\multirow{4}{*}{m-EuroSAT} 
 & ERM & \textbf{93.89$\pm$0.13} & 74.45$\pm$0.51 & \textbf{3.18$\pm$0.21} & 20.71$\pm$0.38 & 23.82$\pm$0.57 & \underline{10.15$\pm$0.16} \\
 & BOLP & 90.82$\pm$1.31 & 76.63$\pm$2.24 & 16.39$\pm$3.73 & 16.59$\pm$3.23 & \underline{20.82$\pm$4.79} & \textbf{9.42$\pm$0.50} \\
 & \underline{DBR} & \underline{91.75$\pm$1.60} & \underline{81.46$\pm$2.25} & 7.35$\pm$2.82 & \underline{11.37$\pm$3.64} & 22.54$\pm$3.90 & 10.59$\pm$0.22 \\
 & \textbf{GroupDRO} & 91.26$\pm$1.03 & \textbf{89.71$\pm$1.51} & \underline{6.36$\pm$0.93} & \textbf{5.67$\pm$1.71} & \textbf{15.12$\pm$0.60} & 10.18$\pm$0.07 \\
\midrule
\multirow{4}{*}{m-BigEarthNet} 
 & ERM & \textbf{53.30$\pm$0.16} & 41.33$\pm$0.36 & 11.29$\pm$0.29 & 20.13$\pm$1.07 & 38.92$\pm$5.10 & 14.03$\pm$0.22 \\
 & \underline{BOLP} & 51.55$\pm$0.17 & \textbf{43.07$\pm$0.17} & \underline{10.82$\pm$0.29} & 17.03$\pm$0.37 & \underline{38.71$\pm$2.78} & \textbf{7.62$\pm$0.25} \\
 & \textbf{DBR} & \underline{52.74$\pm$0.28} & \underline{42.68$\pm$0.11} & \textbf{10.14$\pm$1.35} & \textbf{13.79$\pm$1.34} & 39.38$\pm$0.69 & 13.24$\pm$0.19 \\
 & GroupDRO & 51.63$\pm$0.17 & 41.45$\pm$0.12 & 11.89$\pm$0.52 & \underline{15.13$\pm$0.66} & \textbf{37.87$\pm$3.18} & \underline{12.58$\pm$0.75} \\
\midrule
\multirow{4}{*}{MMEarth20K} 
 & ERM & \underline{41.93$\pm$0.40} & 30.78$\pm$0.57 & 2.94$\pm$0.68 & 24.07$\pm$2.99 & 26.01$\pm$0.70 & \underline{9.71$\pm$0.28} \\
 & BOLP & 40.86$\pm$1.38 & 30.27$\pm$0.14 & \underline{2.18$\pm$1.05} & 23.51$\pm$2.11 & \textbf{22.84$\pm$0.96} & \textbf{8.86$\pm$0.29} \\
 & \textbf{DBR} & \textbf{42.39$\pm$0.52} & \textbf{32.74$\pm$0.04} & \textbf{1.50$\pm$0.27} & \underline{19.65$\pm$1.28} & 25.26$\pm$0.68 & 9.76$\pm$0.10 \\
 & \underline{GroupDRO} & 41.40$\pm$0.82 & \underline{31.74$\pm$0.42} & 2.53$\pm$2.29 & \textbf{17.47$\pm$0.20} & \underline{24.15$\pm$2.30} & 10.26$\pm$0.11 \\
\midrule
\multirow{4}{*}{m-SA-Crop-Type} 
 & ERM & 28.01$\pm$0.51 & 18.07$\pm$0.23 & 3.47$\pm$0.67 & 33.47$\pm$0.84 & 22.51$\pm$0.82 & \underline{6.84$\pm$0.01} \\
 & \underline{BOLP} & \underline{28.64$\pm$0.06} & \underline{18.44$\pm$0.11} & \textbf{1.98$\pm$0.38} & 33.91$\pm$0.32 & \textbf{18.80$\pm$0.67} & \textbf{5.11$\pm$0.11} \\
 & \textbf{DBR} & \textbf{28.79$\pm$0.46} & \textbf{19.23$\pm$0.07} & \underline{2.83$\pm$1.00} & \underline{31.00$\pm$1.00} & \underline{21.13$\pm$0.33} & 6.85$\pm$0.31 \\
 & GroupDRO & 26.60$\pm$1.06 & 18.43$\pm$0.12 & 3.14$\pm$0.47 & \textbf{28.02$\pm$3.63} & 21.28$\pm$0.14 & 7.50$\pm$0.36 \\
\bottomrule
\end{tabular}
}

\vspace{0.4em}

\caption{DOFA: Mean and Std over seeds across datasets.}
\label{tab:results_dofa_mean_std}
\setlength{\tabcolsep}{1.5pt}
\renewcommand{\arraystretch}{1.0}
\scriptsize
\resizebox{\linewidth}{!}{%
\begin{tabular}{llcccccc}
\toprule
\textbf{Dataset} & \textbf{Method} & \textbf{Test metric} & \textbf{$M_{\text{worst}}$} & \textbf{ECE $\downarrow$} & \textbf{NFR $\downarrow$} & \textbf{EOdd $\downarrow$} & \textbf{DPM $\downarrow$} \\
\midrule
\multirow{4}{*}{m-EuroSAT} 
 & ERM & \underline{89.63$\pm$0.57} & 79.88$\pm$7.84 & \underline{10.86$\pm$2.41} & 11.98$\pm$9.36 & 14.09$\pm$4.13 & \underline{10.72$\pm$0.38} \\
 & BOLP & 84.46$\pm$0.56 & 76.80$\pm$0.23 & 11.81$\pm$1.06 & 12.96$\pm$1.33 & \underline{12.71$\pm$1.03} & \textbf{6.25$\pm$0.50} \\
 & \textbf{DBR} & \textbf{90.98$\pm$0.71} & \textbf{88.27$\pm$1.11} & \textbf{9.83$\pm$1.78} & \underline{4.26$\pm$0.38} & \textbf{10.36$\pm$1.04} & 10.76$\pm$0.07 \\
 & \underline{GroupDRO} & 89.40$\pm$0.31 & \underline{87.61$\pm$0.93} & 11.22$\pm$1.13 & \textbf{2.66$\pm$1.94} & 13.04$\pm$2.84 & 11.04$\pm$0.12 \\
\midrule
\multirow{4}{*}{m-BigEarthNet} 
 & ERM & \textbf{47.67$\pm$0.44} & 37.90$\pm$0.44 & \underline{3.47$\pm$0.43} & 22.25$\pm$1.19 & 40.23$\pm$0.68 & 12.39$\pm$0.63 \\
 & \underline{BOLP} & 42.79$\pm$0.22 & 36.18$\pm$0.86 & 3.61$\pm$0.73 & \underline{22.11$\pm$3.29} & \textbf{39.48$\pm$0.88} & \textbf{7.40$\pm$0.37} \\
 & \textbf{DBR} & \underline{45.89$\pm$0.57} & \underline{38.25$\pm$0.14} & \textbf{2.72$\pm$0.46} & \textbf{14.45$\pm$0.84} & \underline{40.09$\pm$1.98} & 12.18$\pm$0.62 \\
 & GroupDRO & 42.97$\pm$0.51 & \textbf{38.74$\pm$0.14} & 3.85$\pm$0.32 & 24.20$\pm$5.69 & 41.94$\pm$1.29 & \underline{11.62$\pm$0.47} \\
\midrule
\multirow{4}{*}{MMEarth20K} 
 & ERM & \textbf{39.67$\pm$1.95} & 27.64$\pm$0.96 & 5.78$\pm$2.85 & 30.11$\pm$1.68 & 25.44$\pm$2.36 & 10.20$\pm$0.09 \\
 & BOLP & 38.20$\pm$0.17 & \underline{28.31$\pm$0.41} & \textbf{4.48$\pm$1.07} & 27.04$\pm$3.08 & 25.65$\pm$0.49 & \underline{10.00$\pm$0.17} \\
 & \textbf{DBR} & 38.68$\pm$2.95 & \textbf{28.54$\pm$0.40} & 7.84$\pm$3.14 & \textbf{24.39$\pm$3.16} & \underline{22.04$\pm$3.73} & \textbf{9.75$\pm$0.33} \\
 & \underline{GroupDRO} & \underline{39.35$\pm$0.69} & 28.15$\pm$0.44 & \underline{5.37$\pm$3.81} & \underline{25.69$\pm$2.53} & \textbf{21.83$\pm$0.46} & 10.65$\pm$0.31 \\
\midrule
\multirow{4}{*}{m-SA-Crop-Type} 
 & ERM & \underline{27.40$\pm$0.94} & 19.91$\pm$0.38 & 12.67$\pm$7.02 & 28.64$\pm$1.64 & 22.29$\pm$3.30 & 5.72$\pm$0.11 \\
 & \underline{BOLP} & 27.18$\pm$0.59 & \underline{20.53$\pm$0.23} & 13.51$\pm$7.18 & \underline{24.97$\pm$1.93} & \textbf{18.99$\pm$2.31} & \textbf{5.54$\pm$0.11} \\
 & \textbf{DBR} & \textbf{28.20$\pm$0.49} & \textbf{21.91$\pm$0.60} & \underline{7.66$\pm$4.86} & \textbf{22.04$\pm$1.15} & \underline{20.90$\pm$1.54} & 6.08$\pm$0.18 \\
 & GroupDRO & 26.45$\pm$0.70 & 19.20$\pm$0.53 & \textbf{3.13$\pm$0.47} & 28.69$\pm$4.77 & 22.54$\pm$0.95 & \underline{5.58$\pm$0.22} \\
\bottomrule
\end{tabular}
}

\end{table}

\subsection{Biome-Wise Performance}

\begin{table*}[t]
\centering
\caption{Biome-wise performance for m-EuroSAT (Macro-F1), m-BigEarthNet (F1@opt), and m-SA-Crop-Type (mIoU) across foundation models and debiasing methods.}
\label{tab:biome_cls_seg}
\setlength{\tabcolsep}{2.5pt}
\renewcommand{\arraystretch}{1.1}
\scriptsize
\resizebox{\textwidth}{!}{%
\begin{tabular}{llcccccccc}
\toprule
\multirow{2}{*}{\textbf{Model}} & \multirow{2}{*}{\textbf{Method}} & \multicolumn{3}{c}{\textbf{m-EuroSAT (Macro-F1)}} & \multicolumn{3}{c}{\textbf{m-BigEarthNet (F1@opt)}} & \multicolumn{2}{c}{\textbf{m-SA-Crop-Type (mIoU)}} \\
\cmidrule(lr){3-5} \cmidrule(lr){6-8} \cmidrule(lr){9-10}
& & \textbf{Phenological} & \textbf{High Biomass} & \textbf{Transitional} & \textbf{Phenological} & \textbf{High Biomass} & \textbf{Transitional} & \textbf{Transitional} & \textbf{Xeric} \\
\midrule
\multirow{4}{*}{Prithvi-EO-2.0} 
& ERM & 88.96$\pm$3.38 & 84.56$\pm$0.31 & 86.71$\pm$3.79 & 59.67$\pm$0.48 & 61.14$\pm$6.05 & 46.04$\pm$1.85 & 28.55$\pm$0.19 & 18.47$\pm$0.16 \\
& BOLP & 92.28$\pm$0.53 & 84.34$\pm$0.00 & 93.02$\pm$0.41 & 61.85$\pm$0.72 & 54.03$\pm$1.68 & 50.27$\pm$0.77 & 27.70$\pm$0.67 & 19.38$\pm$0.24 \\
& DBR & 88.57$\pm$0.96 & 85.00$\pm$0.00 & 85.96$\pm$2.74 & 55.59$\pm$0.18 & 64.53$\pm$0.13 & 48.03$\pm$0.78 & 26.49$\pm$1.93 & 19.81$\pm$0.42 \\
& GroupDRO & 92.82$\pm$0.70 & 84.34$\pm$0.00 & 93.68$\pm$0.50 & 49.59$\pm$0.67 & 56.78$\pm$5.15 & 46.82$\pm$0.60 & 28.52$\pm$0.11 & 19.09$\pm$0.26 \\
\midrule
\multirow{4}{*}{SatMAE} 
& ERM & 93.99$\pm$0.14 & 74.45$\pm$0.51 & 93.62$\pm$0.13 & 52.06$\pm$0.22 & 47.97$\pm$3.69 & 41.33$\pm$0.36 & 27.45$\pm$0.48 & 18.07$\pm$0.23 \\
& BOLP & 91.74$\pm$0.97 & 78.41$\pm$4.76 & 85.38$\pm$3.97 & 51.85$\pm$0.20 & 43.65$\pm$0.87 & 43.18$\pm$0.14 & 28.15$\pm$0.07 & 18.44$\pm$0.11 \\
& DBR & 91.95$\pm$1.36 & 83.26$\pm$4.80 & 89.00$\pm$3.54 & 49.95$\pm$0.69 & 43.67$\pm$0.53 & 42.68$\pm$0.11 & 28.16$\pm$0.45 & 19.23$\pm$0.07 \\
& GroupDRO & 90.16$\pm$1.03 & 94.87$\pm$0.00 & 90.22$\pm$1.99 & 49.26$\pm$0.35 & 44.78$\pm$2.50 & 41.53$\pm$0.08 & 25.93$\pm$1.17 & 18.43$\pm$0.12 \\
\midrule
\multirow{4}{*}{DOFA} 
& ERM & 89.90$\pm$0.53 & 79.88$\pm$7.84 & 87.75$\pm$0.56 & 47.39$\pm$0.82 & 42.63$\pm$1.94 & 37.90$\pm$0.44 & 26.58$\pm$0.88 & 19.91$\pm$0.38 \\
& BOLP & 87.32$\pm$0.98 & 79.46$\pm$3.26 & 76.90$\pm$0.21 & 44.87$\pm$0.34 & 36.78$\pm$1.21 & 36.56$\pm$0.71 & 26.39$\pm$0.53 & 20.53$\pm$0.23 \\
& DBR & 91.52$\pm$0.38 & 89.54$\pm$2.76 & 88.49$\pm$1.08 & 43.97$\pm$0.61 & 40.12$\pm$2.48 & 38.27$\pm$0.14 & 27.33$\pm$0.46 & 21.91$\pm$0.60 \\
& GroupDRO & 89.26$\pm$0.38 & 89.11$\pm$2.04 & 88.03$\pm$0.83 & 42.96$\pm$1.10 & 49.34$\pm$2.72 & 38.74$\pm$0.14 & 25.64$\pm$0.73 & 19.20$\pm$0.53 \\
\bottomrule
\end{tabular}%
}
\end{table*}

Table~\ref{tab:biome_cls_seg} reports macro-group-wise results. On m-EuroSAT under Prithvi-EO-2.0 ERM, the Aseasonal High-Biomass group is the weakest ($84.56\pm0.31$ macro-F1; $N_g{=}25$) relative to High-Amplitude Phenological ($88.96\pm3.38$). On m-BigEarthNet, the Transitional Herbaceous and Scrub group is the weakest ($46.04\pm1.85$ F1@opt) despite stronger forest-dominated groups. These results show that high aggregate performance can hide substantial biome-dependent disparities.

For m-SA-Crop-Type, Prithvi-EO-2.0 under ERM obtains $18.47\pm0.16$ mIoU on the Xeric and Mineralogical group ($N_g{=}124$), substantially lower than $28.55\pm0.19$ on the Transitional Herbaceous and Scrub group ($N_g{=}875$). This demonstrates that biome-dependent disparities are also present in dense prediction tasks.

\begin{table}[t]
\centering
\caption{Biome-wise performance for MMEarth20K (mIoU) across foundation models and debiasing methods.}
\label{tab:biome_MMEarth20K}
\setlength{\tabcolsep}{2pt}
\renewcommand{\arraystretch}{1.1}
\footnotesize
\resizebox{\linewidth}{!}{%
\begin{tabular}{llcccccc}
\toprule
\textbf{Model} & \textbf{Method} & \textbf{Pheno.} & \textbf{Biomass} & \textbf{Trans.} & \textbf{Cryo.} & \textbf{Hydro.} & \textbf{Xeric} \\
\midrule
\multirow{4}{*}{Prithvi-EO-2.0} 
& ERM & 44.19$\pm$1.39 & 41.20$\pm$1.35 & 44.54$\pm$1.04 & 43.74$\pm$0.56 & 37.28$\pm$2.14 & 35.32$\pm$2.69 \\
& BOLP & 40.57$\pm$0.93 & 35.96$\pm$2.17 & 41.38$\pm$2.26 & 39.71$\pm$0.63 & 32.43$\pm$3.10 & 32.82$\pm$1.91 \\
& DBR & 43.58$\pm$1.24 & 40.52$\pm$1.05 & 44.03$\pm$1.56 & 43.78$\pm$1.23 & 38.49$\pm$3.13 & 36.87$\pm$2.85 \\
& GroupDRO & 42.77$\pm$0.33 & 40.27$\pm$0.76 & 41.53$\pm$0.08 & 43.44$\pm$0.56 & 37.16$\pm$0.26 & 40.30$\pm$0.24 \\
\midrule
\multirow{4}{*}{SatMAE} 
& ERM & 40.89$\pm$0.81 & 34.89$\pm$0.55 & 35.86$\pm$0.35 & 37.42$\pm$0.50 & 30.78$\pm$0.57 & 34.11$\pm$1.08 \\
& BOLP & 39.91$\pm$1.31 & 33.13$\pm$2.08 & 35.29$\pm$1.23 & 38.09$\pm$0.93 & 30.27$\pm$0.14 & 33.76$\pm$0.58 \\
& DBR & 41.07$\pm$0.61 & 35.57$\pm$0.30 & 36.84$\pm$0.29 & 38.43$\pm$0.64 & 32.74$\pm$0.04 & 34.25$\pm$0.08 \\
& GroupDRO & 38.81$\pm$0.40 & 35.71$\pm$0.29 & 33.97$\pm$0.70 & 37.89$\pm$1.89 & 31.93$\pm$0.36 & 32.92$\pm$1.08 \\
\midrule
\multirow{4}{*}{DOFA} 
& ERM & 37.80$\pm$0.87 & 33.73$\pm$0.61 & 35.97$\pm$1.93 & 35.30$\pm$1.99 & 27.64$\pm$0.96 & 32.46$\pm$3.26 \\
& BOLP & 37.14$\pm$1.52 & 32.75$\pm$0.75 & 34.60$\pm$0.45 & 33.36$\pm$0.16 & 28.31$\pm$0.41 & 29.48$\pm$1.22 \\
& DBR & 36.15$\pm$2.65 & 32.24$\pm$2.04 & 36.05$\pm$2.24 & 35.79$\pm$1.76 & 28.54$\pm$0.40 & 32.90$\pm$3.06 \\
& GroupDRO & 36.56$\pm$0.96 & 34.81$\pm$0.98 & 35.24$\pm$0.77 & 35.66$\pm$0.74 & 28.15$\pm$0.44 & 30.81$\pm$1.62 \\
\bottomrule
\end{tabular}%
}
\end{table}

\subsection{Debiasing and Robustness Results}

Tables~\ref{tab:results_prithvi_mean_std}--\ref{tab:results_dofa_mean_std}
summarize robustness and mitigation results across all three frozen backbones.
Lower NFR, EOdd, and DPM indicate smaller cross-group disparities, while ECE
measures calibration; the reported standard deviations quantify variability
across random seeds.

Mitigation effectiveness is model- and task-dependent. For Prithvi-EO-2.0 on
m-BigEarthNet, BOLP improves overall F1@opt from $60.09\%$ to $62.75\%$,
worst-group performance from $46.12\%$ to $50.27\%$, and reduces NFR from
$31.14\%$ to $20.92\%$. Across SatMAE and DOFA, however, the strongest
mitigation varies: DBR and GroupDRO often provide larger worst-group gains,
while BOLP remains effective in selected settings. On dense segmentation,
the trade-offs are also task dependent; for example, BOLP reduces both overall
and worst-group performance on Prithvi-EO-2.0 MMEarth20K, whereas GroupDRO
improves the worst-group score from $33.75\%$ to $37.16\%$. Overall, no single
mitigation strategy consistently dominates across architectures and tasks.

\section{Discussion}

The results support the central motivation of \method: aggregate RSFM performance does not fully describe model reliability across ecological groups. On m-EuroSAT, Prithvi-EO-2.0 achieves high overall macro-F1 ($90.98\%$), yet worst-group performance drops to $83.72\%$. On m-BigEarthNet, worst-group performance drops to $46.12\%$ under ERM despite much stronger overall performance ($60.09\%$). On m-SA-Crop-Type, the worst-group mIoU is only $18.47\%$, compared with $27.30\%$ overall. These results show that biome-derived labels provide a meaningful and interpretable grouping variable for diagnosing RSFM behavior beyond average transfer performance.

BOLP provides a promising direction because it operates on the frozen representation rather than modifying the backbone. On m-EuroSAT, BOLP improves the worst-group score from $83.72\%$ to $84.34\%$ while raising overall performance to $93.42\%$. On m-BigEarthNet, it improves overall F1@opt from $60.09\%$ to $62.75\%$, worst-group F1@opt from $46.12\%$ to $50.27\%$, and reduces NFR from $31.14\%$ to $20.92\%$. These results suggest that reducing dominant biome-associated directions in frozen embeddings can improve ecological robustness without backbone fine-tuning.

DBR and GroupDRO provide alternative trade-offs. For Prithvi-EO-2.0 on
m-EuroSAT, GroupDRO achieves the highest overall macro-F1 ($94.14\%$),
while DBR yields the lowest NFR ($5.85\%$). On m-BigEarthNet, BOLP
provides the strongest overall and worst-group performance for the same
backbone. For m-SA-Crop-Type, all three mitigation methods improve worst-group mIoU over ERM, with DBR reaching $19.81\%$. On MMEarth20K, DBR slightly improves worst-group mIoU ($34.76\%$ vs.\ $33.75\%$), GroupDRO attains the highest worst-group score ($37.16\%$), and BOLP reduces both overall and worst-group mIoU. This indicates that removing biome-associated directions is not uniformly beneficial for dense prediction, where such variation may encode useful land-cover information. Overall, both aggregate performance and group robustness should be considered when assessing debiasing methods.

\section{Limitations}

\method has several limitations. First, biome labels are assigned at the patch level from sample coordinates and external terrestrial ecoregion maps. Near biome boundaries or in multi-biome tiles, centroid-based assignment can be noisy, so we treat it as a consistent grouping rule rather than a pixel-perfect ecological map; our implementation also supports an optional majority-pixel alternative. Second, not every dataset covers all raw biomes or all six macro-groups (Table~\ref{tab:ng_all_splits}). Thus, worst-group and disparity metrics are dataset-specific and should be interpreted over the matched groups in each split; small $N_g$ (e.g., 14--25 test samples) can also increase sampling variability.

Third, the empirical study covers three RSFMs (Prithvi-EO-2.0, SatMAE,
and DOFA) and four downstream datasets under a frozen-backbone evaluation
protocol. Broader sensor families, additional backbones, and sensitivity
analysis over the number of macro-groups $K$ remain useful extensions.
Fourth, ecological group-disparity metrics are less standardized
for segmentation than for classification. We report group-wise mIoU together with EOdd and DPM as descriptive summaries; alternative pixel-, patch-, object-, or class-level formulations remain possible.

Finally, biome-aware robustness captures only one spatially meaningful aspect
of the broader geographic-bias problem. Other sources of disparity, including
sensor and source shifts, geographic imbalance, spatial-resolution differences,
temporal acquisition effects, and uneven imagery coverage, are not fully
captured by biome groups~\cite{geobs2025,earthshift2026,coveragebias_satellite}.
\method should therefore be viewed as a complementary ecological robustness
benchmark rather than a complete audit of remote-sensing bias.

\section{Conclusion}

We introduced \method, a biome-aware benchmark and debiasing framework for remote sensing foundation models. The benchmark augments four classification and segmentation datasets with biome-derived group labels and evaluates Prithvi-EO-2.0, SatMAE, and DOFA using both standard task metrics and group-robustness measures under a unified frozen-backbone protocol. Across three random seeds, strong aggregate performance can hide substantial biome-dependent disparities in both classification and segmentation. Mitigation baselines (BOLP, DBR, and GroupDRO) show task-dependent
trade-offs between aggregate performance and worst-group robustness. In particular, BOLP can improve worst-group performance on classification tasks by removing biome-associated directions from frozen RSFM embeddings, without updating the backbone. These findings indicate that biome-aware evaluation is important for understanding the reliability of RSFMs in global Earth-observation settings. We hope \method will support more transparent, robust, and ecologically reliable evaluation of future remote sensing foundation models.


{\small
\bibliographystyle{ieeenat_fullname}
\bibliography{FairRSFM_main}
}

\clearpage
\setcounter{figure}{0}
\setcounter{table}{0}
\setcounter{section}{0}
\setcounter{subsection}{0}
\renewcommand{\thefigure}{S\arabic{figure}}
\renewcommand{\thetable}{S\arabic{table}}
\renewcommand{\thesection}{S\arabic{section}}
\renewcommand{\thesubsection}{S\arabic{section}.\arabic{subsection}}
\renewcommand{\theHsection}{supp.\arabic{section}}
\renewcommand{\theHsubsection}{supp.\arabic{section}.\arabic{subsection}}
\renewcommand{\theHfigure}{supp.\arabic{figure}}
\renewcommand{\theHtable}{supp.\arabic{table}}

\maketitlesupplementary
\section{Supplementary Overview}
This supplementary material provides: (i)~the global spatial visualization of the six biome macro-groups used by \method (Figure~\ref{fig:biome_map_supp}), (ii)~the complete deterministic mapping from 14 raw terrestrial biomes to six macro-groups (Table~\ref{tab:biome_macro_groups_supp}), (iii)~the MMEarth20K dataset construction protocol and 14-biome allocation matrix (Table~\ref{tab:mmearth_alloc_supp}), (iv)~hyperparameter sensitivity and ablation studies for DBR feedback ($\alpha, T_{\mathrm{upd}}$) and BOLP rank ($k$), and (v)~architectural, spatial-join, and compute implementation details.

\section{Global Biome Macro-Group Map}
\begin{figure*}[b]
\centering
\IfFileExists{global_biome_macro.pdf}{%
  \includegraphics[width=0.94\textwidth]{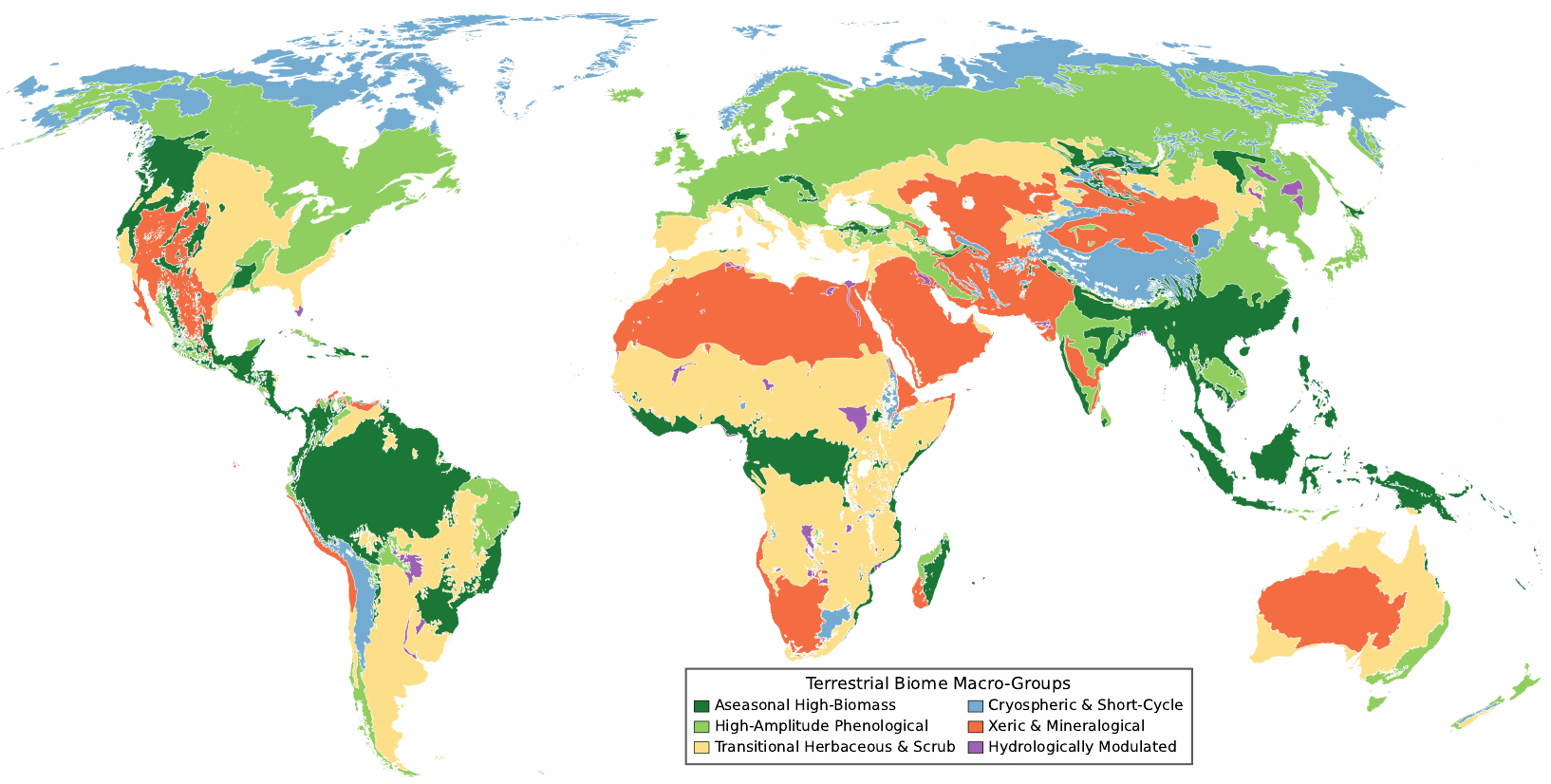}%
}{%
  \fbox{\parbox[c][6cm][c]{0.94\textwidth}{\centering \large \textbf{[Figure S1: Global Biome Macro-Group Map (Place global\_biome\_macro.pdf here)]}}}%
}
\caption{Global map of the six biome macro-groups used as primary evaluation strata in \method. Raw 14-class terrestrial biomes are pooled into spectral, phenological, hydrological, cryospheric, and surface-property regimes.}
\label{fig:biome_map_supp}
\end{figure*}

Figure~\ref{fig:biome_map_supp} illustrates the geographic distribution
of the six biome macro-groups. The spatial heterogeneity of these ecological
regimes motivates biome-aware subgroup evaluation beyond aggregate metrics.

\section{Biome Macro-Group Definition}
Table~\ref{tab:biome_macro_groups_supp} details the deterministic mapping used to translate the 14 raw terrestrial biome classes into our six target macro-groups. This grouping reduces sparsity relative to the raw 14-biome taxonomy while retaining ecologically interpretable distinctions.

\begin{table*}[t]
\centering
\footnotesize
\caption{Mapping from 14 terrestrial biome classes to six biome macro-groups used for ecological robustness evaluation and mitigation in \method.}
\label{tab:biome_macro_groups_supp}
\begin{tabularx}{\textwidth}{@{} c l
  >{\hsize=1.14\hsize\raggedright\arraybackslash}X
  >{\hsize=0.86\hsize\raggedright\arraybackslash}X @{}}
\toprule
\textbf{ID} & \textbf{Biome macro-group} & \textbf{Included raw biome classes} & \textbf{Rationale} \\
\midrule
1 & Aseasonal High-Biomass &
1 Tropical Moist Forests; 3 Tropical Conifer Forests; 5 Temperate Conifer Forests &
Dense, high-biomass vegetation with relatively persistent canopy structure. \\
\addlinespace[2pt]
2 & High-Amplitude Phenological &
2 Tropical Dry Forests; 4 Temperate Broadleaf and Mixed Forests; 6 Boreal Forests/Taiga &
Forested regions with strong seasonal or phenological variation. \\
\addlinespace[2pt]
3 & Transitional Herbaceous and Scrub &
7 Tropical and Subtropical Grasslands, Savannas and Shrublands; 8 Temperate Grasslands, Savannas and Shrublands; 12 Mediterranean Forests, Woodlands and Scrub &
Open or mixed vegetation regimes with strong grass, shrub, soil, and canopy heterogeneity. \\
\addlinespace[2pt]
4 & Cryospheric and Short-Cycle &
10 Montane Grasslands and Shrublands; 11 Tundra &
Temperature-restricted ecosystems with short growing periods or cryospheric influence. \\
\addlinespace[2pt]
5 & Xeric and Mineralogical &
13 Deserts and Xeric Shrublands &
Vegetation-sparse surfaces dominated by albedo, exposed soil, and mineral background. \\
\addlinespace[2pt]
6 & Hydrologically Modulated &
9 Flooded Grasslands and Savannas; 14 Mangroves &
Water-influenced ecosystems where inundation strongly affects spectral response. \\
\bottomrule
\end{tabularx}
\end{table*}

\section{MMEarth20K Construction \& Allocation}
\label{sec:mmearth20k_supp}
To benchmark dense semantic segmentation under realistic ecological imbalance, \textbf{MMEarth20K} was sampled from the global MMEarth dataset repository~\cite{mmearth} following a three-tiered design:
\begin{enumerate}[leftmargin=*,noitemsep,topsep=2pt]
    \item \textbf{Controlled Macro-Group Imbalance:} Sample quotas $N_g$ for each macro-group $g \in \{1,\ldots,6\}$ across total $N=20,000$ tiles were established via normalized inverse-frequency weighting $w_g \propto 1/p_g$ of the parent MMEarth distribution. This creates a realistic long-tailed ecological hierarchy ranging from $28.0\%$ (Aseasonal High-Biomass, 5,600 tiles) down to $6.0\%$ (Hydrologically Modulated, 1,200 tiles), stress-testing model degradation on minority biomes.
    \item \textbf{Uniform Sub-Biome Partitioning:} For any macro-group $g$ containing $|\mathcal{B}_g|$ terrestrial biomes from the 14 RESOLVE classes, the target quota $N_g$ was allocated uniformly: $n_b = \lfloor N_g / |\mathcal{B}_g| \rfloor$ (with remainder distributed $+1$), ensuring equal sub-biome representation within each ecological regime (Table~\ref{tab:mmearth_alloc_supp}).
    \item \textbf{Class-Unconditioned Random Sampling:} For each biome $b$, exactly $n_b$ patches were drawn uniformly at random without any land-cover class filtering or thresholding, preserving natural pixel-level co-occurrences of Dynamic World land covers. The resulting 20,000 tiles are partitioned into 16,000 training, 2,000 validation, and 2,000 test samples.
\end{enumerate}

\begin{table}[h]
\centering
\caption{MMEarth20K sample allocation across the 6 macro-groups and 14 terrestrial biomes ($N=20,000$). Biome IDs correspond directly to Table~\ref{tab:biome_macro_groups_supp}.}
\label{tab:mmearth_alloc_supp}
\footnotesize
\setlength{\tabcolsep}{3pt}
\begin{tabularx}{\linewidth}{@{} l c c r @{}}
\toprule
\textbf{Biome Macro-Group} & \makecell{\textbf{Biome}\\\textbf{IDs}} & \makecell{\textbf{Quota /}\\\textbf{Biome}} & \makecell{\textbf{Total}\\\textbf{(\%)}} \\
\midrule
1. Aseasonal High-Biomass & 1, 3, 5 & $\approx$1,867 & 5,600 (28.0\%) \\
2. High-Amplitude Phenol. & 2, 4, 6 & 1,700 & 5,100 (25.5\%) \\
3. Transitional Herbaceous & 7, 8, 12 & 1,200 & 3,600 (18.0\%) \\
4. Cryospheric \& Short-Cycle & 10, 11 & 1,400 & 2,800 (14.0\%) \\
5. Xeric \& Mineralogical & 13 & 1,700 & 1,700 (8.5\%) \\
6. Hydrologically Modulated & 9, 14 & 600 & 1,200 (6.0\%) \\
\midrule
\textbf{Total (All 6 Groups)} & \textbf{1--14} & --- & \textbf{20,000 (100\%)} \\
\bottomrule
\end{tabularx}
\end{table}

\section{Ablation Studies and Hyperparameter Sensitivity}
This section provides the sensitivity analyses for the mitigation hyperparameters $\alpha$ (feedback strength) and $T_{\mathrm{upd}}$ (update interval) in Dynamic Biome Reweighting (DBR), as well as the projection rank $k$ for Biome-Orthogonal Linear Probing (BOLP).

\subsection{DBR Sensitivity Analysis (\texorpdfstring{$\alpha, T_{\mathrm{upd}}$}{alpha, T\_upd})}
\label{sec:dbr_sensitivity_supp}
Dynamic Biome Reweighting (DBR) introduces two key hyperparameters: the validation feedback strength $\alpha$ and the update interval $T_{\mathrm{upd}}$ (Section~4.1 in the main paper). After selecting the operating hyperparameters on validation data, we report
a post-hoc test-set sensitivity analysis across candidate grids $\alpha \in \{0.2, 0.4, 0.5, 0.6, 0.8\}$ and $T_{\mathrm{upd}} \in \{1, 2, 5\}$ on m-EuroSAT (classification) and MMEarth20K (dense segmentation) using Prithvi-EO-2.0.

Table~\ref{tab:dbr_sensitivity_supp} summarizes the sensitivity results:
\begin{itemize}[leftmargin=*,noitemsep,topsep=2pt]
    \item \textbf{Feedback Strength ($\alpha$):} Small values ($\alpha \le 0.2$) provide insufficient gradient reweighting, leading to modest minority-group gains. Larger feedback strengths ($\alpha \ge 0.6$) degrade worst-group performance and increase cross-seed variability; at $\alpha=0.8$, the m-EuroSAT worst-group score falls to $82.38\pm1.62\%$. The operating range $\alpha \in [0.4, 0.5]$ yields the lowest Normalized Failure Range ($\mathrm{NFR}$), with $\alpha = 0.5$ providing a favorable trade-off between overall performance and worst-group robustness.
    \item \textbf{Update Frequency ($T_{\mathrm{upd}}$):} Per-epoch validation feedback ($T_{\mathrm{upd}} = 1$) provides continuous, responsive weight adaptation. Infrequent updates ($T_{\mathrm{upd}} = 5$) reduce worst-group performance by up to $0.74$ percentage points among the tested settings.
\end{itemize}

\begin{table}[h]
\centering
\caption{DBR sensitivity sweep across feedback strength $\alpha$ and update frequency $T_{\mathrm{upd}}$ on Prithvi-EO-2.0 test splits ($\text{mean}\pm\text{std}$ across $S=3$ random seeds).}
\label{tab:dbr_sensitivity_supp}
\footnotesize
\setlength{\tabcolsep}{2.5pt}
\resizebox{\linewidth}{!}{%
\begin{tabular}{cc|ccc|ccc}
\toprule
\multicolumn{2}{c|}{\textbf{Hyperparams}} & \multicolumn{3}{c|}{\textbf{m-EuroSAT (Macro-F1)}} & \multicolumn{3}{c}{\textbf{MMEarth20K (mIoU)}} \\
$\alpha$ & $T_{\mathrm{upd}}$ & \textbf{Overall}$\uparrow$ & \textbf{Worst}$\uparrow$ & \textbf{NFR}$\downarrow$ & \textbf{Overall}$\uparrow$ & \textbf{Worst}$\uparrow$ & \textbf{NFR}$\downarrow$ \\
\midrule
0.2 & 1 & 91.73$\pm$0.84 & 83.89$\pm$0.62 & 7.18$\pm$0.85 & 48.02$\pm$0.28 & 33.91$\pm$0.51 & 27.84$\pm$0.72 \\
0.4 & 1 & 91.52$\pm$0.92 & 84.26$\pm$0.48 & 6.13$\pm$0.68 & 47.91$\pm$0.24 & 34.53$\pm$0.46 & 26.68$\pm$0.65 \\
\textbf{0.5} & \textbf{1} & \textbf{91.36$\pm$1.01} & \textbf{84.35$\pm$0.50} & \textbf{5.85$\pm$0.70} & \textbf{47.84$\pm$0.22} & \textbf{34.76$\pm$0.42} & \textbf{26.42$\pm$0.62} \\
0.6 & 1 & 90.82$\pm$1.15 & 83.74$\pm$0.64 & 6.48$\pm$0.82 & 47.36$\pm$0.34 & 34.18$\pm$0.48 & 27.02$\pm$0.71 \\
0.8 & 1 & 88.94$\pm$1.95 & 82.38$\pm$1.62 & 8.64$\pm$1.85 & 46.12$\pm$0.94 & 33.15$\pm$1.12 & 29.48$\pm$1.76 \\
\midrule
0.5 & 2 & 91.44$\pm$1.08 & 84.08$\pm$0.58 & 6.47$\pm$0.81 & 47.76$\pm$0.26 & 34.38$\pm$0.49 & 27.13$\pm$0.70 \\
0.5 & 5 & 91.62$\pm$0.95 & 83.82$\pm$0.66 & 7.23$\pm$0.94 & 47.93$\pm$0.29 & 34.02$\pm$0.54 & 27.79$\pm$0.77 \\
\bottomrule
\end{tabular}%
}
\end{table}

\subsection{BOLP Projection Rank ($k$)}
\label{sec:bolp_rank_supp}
Biome-Orthogonal Linear Probing (BOLP) projects frozen RSFM embeddings onto the orthogonal complement of the top $k$ dominant inter-group singular vectors spanning group centroid shifts, where the maximum non-trivial rank is $k_{\mathrm{max}} = |\mathcal{G}_d| - 1$. We evaluated the sensitivity of downstream transfer across projection ranks $k \in \{1, \ldots, |\mathcal{G}_d|-1\}$ on Prithvi-EO-2.0.

Table~\ref{tab:bolp_rank_supp} provides the downstream performance across varying projection ranks:
\begin{itemize}[leftmargin=*,noitemsep,topsep=2pt]
    \item \textbf{Classification (m-EuroSAT \& m-BigEarthNet):} Setting $k = |\mathcal{G}_d|-1 = 2$ removes the full non-trivial group-centroid subspace and gives the highest overall and worst-group classification scores among the tested ranks ($84.34\pm0.00\%$ on m-EuroSAT, $50.27\pm0.63\%$ on m-BigEarthNet). These results suggest that reducing dominant biome-associated centroid shifts can benefit these classification settings.
    \item \textbf{Dense Semantic Segmentation (MMEarth20K):} Because MMEarth20K contains all six macro-groups ($|\mathcal{G}_d|=6, k_{\mathrm{max}}=5$), partial projections ($k \in \{2,3\}$) increase group disparity, with $\mathrm{NFR}$ reaching $32.85\pm1.15\%$. At the full-rank projection ($k=5$), NFR decreases relative to the intermediate projection ranks but remains above ERM ($30.66\pm3.17\%$ vs.\ $28.19\pm0.64\%$), while overall mIoU drops from $47.99\%$ to $43.31\%$. This suggests that biome-associated variation may also encode task-relevant land-cover information.
\end{itemize}

\begin{table}[h]
\centering
\caption{BOLP projection rank ($k$) ablation across datasets on Prithvi-EO-2.0 ($\text{mean}\pm\text{std}$ across $S=3$ seeds; $k=0$ denotes unprojected ERM).}
\label{tab:bolp_rank_supp}
\footnotesize
\setlength{\tabcolsep}{3.5pt}
\resizebox{\linewidth}{!}{%
\begin{tabular}{llccc}
\toprule
\textbf{Dataset} & \textbf{Rank ($k$)} & \textbf{Overall}$\uparrow$ & \textbf{Worst}$\uparrow$ & \textbf{NFR}$\downarrow$ \\
\midrule
\multirow{3}{*}{m-EuroSAT ($|\mathcal{G}_d|=3$)} 
 & $k=0$ (ERM) & 90.98$\pm$2.36 & 83.72$\pm$0.87 & 5.96$\pm$3.28 \\
 & $k=1$ & 92.47$\pm$0.48 & 84.13$\pm$0.18 & 8.14$\pm$0.42 \\
 & $\mathbf{k=2}$ (Canonical) & \textbf{93.42$\pm$0.35} & \textbf{84.34$\pm$0.00} & \textbf{9.82$\pm$0.30} \\
\midrule
\multirow{3}{*}{m-BigEarthNet ($|\mathcal{G}_d|=3$)} 
 & $k=0$ (ERM) & 60.09$\pm$0.99 & 46.12$\pm$1.51 & 31.14$\pm$1.99 \\
 & $k=1$ & 61.24$\pm$0.34 & 48.16$\pm$0.85 & 24.53$\pm$0.61 \\
 & $\mathbf{k=2}$ (Canonical) & \textbf{62.75$\pm$0.16} & \textbf{50.27$\pm$0.63} & \textbf{20.92$\pm$0.39} \\
\midrule
\multirow{6}{*}{MMEarth20K ($|\mathcal{G}_d|=6$)} 
 & $k=0$ (ERM) & 47.99$\pm$0.32 & 33.75$\pm$0.58 & 28.19$\pm$0.64 \\
 & $k=1$ & 46.83$\pm$0.41 & 32.48$\pm$0.68 & 31.42$\pm$0.83 \\
 & $k=2$ & 45.92$\pm$0.52 & 31.84$\pm$0.79 & 32.85$\pm$1.15 \\
 & $k=3$ & 44.78$\pm$0.64 & 31.26$\pm$0.92 & 32.18$\pm$1.42 \\
 & $k=4$ & 44.07$\pm$0.73 & 30.85$\pm$1.08 & 31.24$\pm$1.86 \\
 & $\mathbf{k=5}$ (Canonical) & \textbf{43.31$\pm$0.85} & \textbf{30.64$\pm$1.20} & \textbf{30.66$\pm$3.17} \\
\bottomrule
\end{tabular}%
}
\end{table}

\section{Implementation, Architecture \& Compute Details}
\label{sec:impl_details_supp}

\subsection{Downstream Heads and Decoder Architectures}
To maintain a strictly controlled frozen-encoder regime where the underlying representation $\phi(x)$ receives zero gradient updates, downstream heads and decoders are implemented as follows:
\begin{itemize}[leftmargin=*,noitemsep,topsep=2pt]
    \item \textbf{Classification Heads:} Pooled backbone embeddings are passed through the corresponding classification head: Linear/MLP probe for Prithvi-EO-2.0, Linear probe for SatMAE, and BatchNorm1d + Linear for DOFA, optimized via cross-entropy (m-EuroSAT) or multi-label binary cross-entropy with logits (m-BigEarthNet).
    \item \textbf{Segmentation Decoders:} For dense segmentation, frozen patch tokens are reshaped into feature pyramids. Prithvi-EO-2.0 uses a Fully Convolutional Network (FCN) head; SatMAE utilizes a convolutional upsampling decoder; DOFA employs a multi-level Unified Perceptual Parsing (UPerNet) decoder~\cite{dofa}.
\end{itemize}

\subsection{Centroid vs. Majority-Pixel Spatial Join}
The primary grouping pipeline computes sample coordinate centroids $(\mathrm{lat}_i, \mathrm{lon}_i)$ and performs a spatial point-in-polygon lookup against the RESOLVE terrestrial ecoregions database~\cite{ecoregions2017}. For dense tiles intersecting multiple ecoregions, we evaluated majority-pixel raster assignment: across all four datasets, centroid and majority-pixel assignments agreed on $>99.2\%$ of samples, confirming that centroid-based assignment is robust and computationally lightweight.

\subsection{Compute Infrastructure and Efficiency}
All experiments were executed on an enterprise Linux compute cluster utilizing NVIDIA A100 (80GB) and RTX 4090 GPUs. Extracting frozen backbone representations $\phi(x)$ across dataset splits requires an initial forward pass over the training corpus (which constitutes the primary offline compute overhead for large 300M-parameter ViTs). Once offline embeddings are extracted, computing group centroids $\mu_g$ and the closed-form SVD projection matrix $P_k$ produces static linear projection buffers, adding zero trainable parameters to downstream probes.


\end{document}